# ON-BOARD ANOMALY DETECTION FOR EFFICIENT MARINE ENVIRONMENTAL MONITORING*

**Thomas Goudemant, Clotilde Szywala, Benjamin Francesconi, Michelle Aubrun, Yves Bobichon, Marjorie Bellizzi, Adrien Girard**

***Institut de Recherche Technologique Saint Exupéry***
*240 Rue Evariste Galois*
*06410 Biot, France*
*Email: firstname.lastname@irt-saintexupery.com*

## ABSTRACT

Marine ecosystems are impacted by various threats such as oil spills, algal blooms, and sediment floods, which disrupt habitats, wildlife, and human activities. Advances in satellite imagery and Artificial Intelligence (AI) have enhanced our capabilities for early detection and mitigation of such hazards. In this paper, we propose a marine event detection pipeline for Earth observation satellites equipped with multi- or hyperspectral sensors. Our approach includes a self-supervised neural network encoder that compresses satellite images into a reduced latent space, enabling efficient onboard processing. A machine learning anomaly detection model identifies deviations from normal sea patterns to detect environmental anomalies. We compare its performance against traditional algorithms such as Isolation Forest, One-Class Support Vector Machine and Local Outlier Factors. Our lightweight, resource-efficient pipeline is optimized for deployment on satellites with limited computational resources, ranging from embedded CPUs to AI hardware accelerators. By prioritizing the transmission of critical information, our solution enhances system responsiveness and optimizes satellite communication bandwidth. Demonstrated through current integration across multiple missions, including European Space Agency's (ESA) Φsat-2 mission and Microsoft/Thales Alenia Space IMAGIN-e mission, our pipeline aims to improve marine environmental monitoring by providing timely alerts and efficient data reduction.

## 1. INTRODUCTION

Oceans support diverse marine species and play a key role in Earth's biodiversity. Technologies like satellite imagery and Artificial Intelligence (AI) are essential for identifying and mitigating threats such as harmful algal blooms, oil spills, and sediment floods. Early detection helps limit ecological damage, protect fisheries, and aid cleanup operations [1], [2].

Current research often targets specific threats through methods like water quality assessment [3], [4], monitoring suspended solids [5], [6], detecting oil spills [7], and identifying plastic debris [8]. These methods rely on combining spectral bands, which can be challenging to adapt across different satellites missions or geographic areas. Traditional machine learning approaches [9], [10], [11] also struggle with generalization. While deep learning shows state-of-the-art results, supervised techniques [12], [10], [11] require large labelled datasets and may not generalize well to other use cases. Self-supervised [9], [13], [11] or unsupervised approaches based on image reconstruction [9], [13], [10], [11] offer better generalization but remain too computationally heavy for the constrained hardware of satellite platforms.

Our approach innovatively combines the strengths of deep learning with a self-supervised encoder, enabling efficient compression while employing a generic anomaly detection strategy. This enables us to identify a wide range of unexpected events, including environmental threats, in marine ecosystems, even when they can't be precisely characterized. We refer to these events as anomalies—any elements that deviate from a defined normality [14], represented by maritime images of unaffected ecosystems. This generic approach reduces the need for extensive annotations and enhances adaptability to different missions and sensors.

Designed for lightweight satellite hardware, our solution is deployable on small satellites and CubeSats. It aligns with the European Space Agency's (ESA) OrbitalAI challenges, which promote edge computing in orbit by selecting innovative AI solutions for two remote sensing missions: Φsat-2 and IMAGIN-e. Φsat-2, a 6U CubeSat in sun-synchronous orbit, is equipped with a camera offering 5-meter spatial resolution and 8 spectral bands in the visible and near-infrared, utilizing the CogniSAT™ processor with an Intel® Movidius™ Myriad™ 2 accelerator. IMAGIN-e, a space edge computing demonstration by Microsoft and Thales Alenia Space hosted on the International Space Station, includes a hyperspectral camera with 50-meter spatial resolution, 50 spectral bands, and a 16-core Cortex-A72 processor.

Edge computing improves responsiveness by processing data directly on board, reducing the time between detection and response, while optimizing satellite bandwidth by prioritizing critical data. A key feature of our solution is its ability to encode data into a highly compressed latent space, enabling near real-time onboard processing of large areas such as the ocean. This compressed data can be transmitted for further ground-based analysis, including more advanced anomaly detection, change detection or image reconstruction. Selected by ESA for Φsat-2 and IMAGIN-e, we are also currently discussing its potential deployment on the Australian Kanyini CubeSat and Irish CogniSAT-6 CubeSat, which are equipped with hyperspectral and multispectral sensors respectively.
The following sections detail our solution: Section 2 describes the architecture, while Section 3 focuses on the datasets and pre-processing steps used for training. Section 4 presents experimental results on anomaly detection and hardware performance, followed by Section 5 discussing real-time applications and potential operational deployment. Finally, Section 6 concludes with key takeaways and future perspectives for onboard marine anomaly detection.

** This paper was presented at the 9th International Workshop on On-Board Payload Data Compression (OBPDC 2024), Gran Canaria, Spain, 2–4 October 2024. Conference version: https://doi.org/10.5281/zenodo.13850918*

## 2. METHODS

In this section, we describe the architecture of our solution and justify the selection of the AI models used.

### 2.1. Architecture Description

The architecture of our solution, presented in Fig. 1, consists of three components: a *Tiler* to divide the image into 32 × 32 pixel patches, an *AI pipeline* to detect and characterize anomalies, and a *Mosaic'er* to visualize the model outputs. The *AI pipeline*, at the core of our solution, operates in three stages with four successive AI models:

1. *Sea/No-Sea Segmentation:* A Non-Linear Support Vector Machine (NLSVM) filters out non-relevant patches, focusing processing on marine areas.
2. *Anomaly Detection:* Potential threats are detected by measuring the deviation of each patch from a learned representation of the normal sea state. The patch data is compressed into a latent space using a neural network encoder trained with the self-supervised learning method SimCLR (Simple Contrastive Learning of Representations) [15], which significantly reduces data volume. The compressed features are then compared to a normal distribution modelled by a Gaussian Mixture Model (GMM) [16], which assigns an anomaly score to each patch. Patches with high anomaly scores are selected for further analysis.
3. *Anomaly Characterization:* For patches exceeding the anomaly threshold, a multi-class NLSVM performs pixel-level segmentation into four categories: oil spill, algal bloom, sediments, and sea, providing detailed information about detected anomalies.

The outputs from each stage are recombined by the *Mosaic'er*, which produces a sea/no-sea segmentation map, an anomaly heatmap, and a pixel-level segmentation map of anomaly types.
This modular solution offers a comprehensive approach and allows multipurpose use for different operational applications, such as:

- *Prioritizing Images:* Images with the highest anomaly scores are downloaded first by the satellite. This allows operators to focus on important data when monitoring vast marine areas, significantly improving the efficiency of data analysis and decision-making processes.
- *Sending Alerts:* If a major incident at sea—such as an oil spill—is detected, an alert is immediately triggered. This enables a rapid response by relevant authorities, mitigating environmental, social, and economic impacts.

In addition to enhancing system reactivity, our approach also focuses on efficient data reduction. By encoding input data into a compact latent space and prioritizing the transmission of anomalous or high-priority patches, the system drastically reduces the volume of data sent to the ground. This ensures that only the most relevant information is transmitted, optimizing the limited satellite communication bandwidth and reducing the latency between detection and response. Together, these features improve the overall effectiveness of satellite operations for marine environmental monitoring.

### 2.2. Architecture Justification

#### *2.2.1. Sea Segmentation and Anomaly Characterization*

For sea segmentation and anomaly characterization, we use NLSVM models. This is the non-linear extension of the Support Vector Machine (SVM) model [17], [18], which consists of reducing a binary classification problem to a hyperplane. In our case, the equation of the hyperplane is a linear combination of features such as spectral bands, weighted band ratios, and weighted band indices. The linear combination parameters (weights of a dense layer) and band weights (non-linear parameters of the NLSVM) are determined through supervised learning.

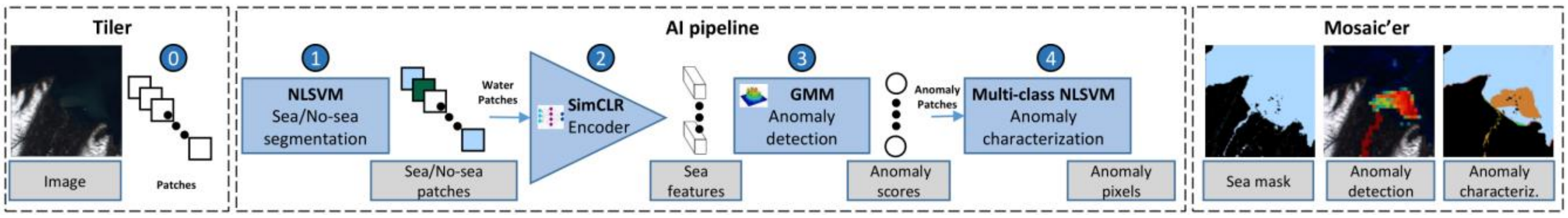

Fig. 1. Architecture of the anomaly detection pipeline: The Tiler (0) splits the input image into patches, which are processed by the AI pipeline. Sea patches are filtered (1), encoded into a latent space (2), and analyzed for anomalies (3). Anomalies are segmented (4) to identify their type. The Mosaic'er then recombines the patches for visualization.

Sea/no-sea segmentation is handled as a binary classification, while anomaly characterization is based on a segmentation of patches into four classes: oil, algae, sediments, and sea (to classify sea pixels without anomalies within the patch). We use three parallel NLSVMs, each focusing on one anomaly type (sea/oil, sea/algae, and sea/sediments), and train them together to classify each pixel. We refer to this model as the *multi-class NLSVM*.

To maximize the performance of our models, we selected features (bands, indices, and ratios) that best discriminate between classes. We analyzed the average spectra and their dispersion for each class, and evaluated the distribution of values for all possible combinations: 2,450 indices and 2,450 ratios. We retained those that offered the highest discrimination between classes. This selection process was validated by comparing our findings with the results reported in relevant scientific literature [19], [20], [21], [22]. Based on this analysis, we selected 7 spectral bands for IMAGIN-e, in line with mission constraints, while using all 8 available bands for Φsat-2.

NLSVMs offer several advantages: they are based on physical features, fast in inference, and easy to implement on embedded systems. Their lightweight structure allows them to run efficiently on hardware CPUs as part of pre-processing (sea segmentation) and post-processing (anomaly characterization), ensuring compatibility with satellite resource constraints.

#### 2.2.2. *Anomaly Detection*

For anomaly detection, we use a self-supervised deep learning approach that combines a lightweight encoder with a GMM. The encoder, trained with the SimCLR method, learns robust data representations without requiring labeled samples. This allows it to be retrained on real satellite data, adapting to new missions and sensors without manual annotation.

The encoder is a compact neural network with four convolutional layers and about one million parameters, much smaller than encoder models like ResNet18 [23], making it highly suitable for deployment on resource-constrained hardware such as the Intel® Movidius Myriad™ accelerators or ARM CPUs. It compresses 32 × 32 multispectral patches into a latent space of 16 values, reducing data dimensionality by over 500 times while retaining essential information. These robust encoded representations capture the key features of entire patches, enabling more advanced processing on the ground.

For anomaly scoring, we selected the GMM, which accurately models the distribution of the latent space from normal sea conditions. By using several dozen Gaussian components, the GMM captures the complexity of normal data, detecting anomalies as deviations from learned patterns. This weakly supervised approach is data-efficient, requiring only tens of thousands of sea patches for training and enabling detection of various anomalies without needing explicit examples. Together, the self-supervised encoder and GMM provide an effective, adaptable solution for anomaly detection on small satellite missions.

## 3. DATABASE

Our pipeline models require datasets tailored to their specific training needs, including marine normality data (without anomalies) and anomaly data. Both datasets are designed to match the sensors on the Φsat-2 and IMAGIN-e missions.

### 3.1. Database Overview

#### *3.1.1. Marine Normality Database*

This database consists of about 60 scenes and cloud masks simulated according to Φsat-2 and IMAGIN-e missions characteristics, provided through ESA's OrbitalAI challenges. The scenes cover diverse environments such as marine areas, land, and coastlines, which we manually labelled with sea/land segmentation masks. IMAGIN-e images are 1,024 × 1,024 pixels with 50 spectral bands (VIS/NIR) at 50 meters resolution, while Φsat-2 images are 4,096 × 4,096 pixels with 8 bands at 5 meters resolution. The encoder was trained in an unsupervised manner on these scenes, while sea patches were later used to train the GMM anomaly detector and sea segmentation masks to train the sea/no-sea NLSVM.

#### *3.1.2. Anomaly Database*

Our anomaly database consists of 31 simulated images representative of Φsat-2 and IMAGIN-e, generated using the simulators described in Section 3.2. These images represent pollution events sourced from scientific literature, expert contributions, and press releases, covering oil spills [24], algal blooms [25], [26], or sediments and plastic [27], [28], [29]. Each scene was manually annotated with masks for land, sea, cloud, and anomalies. Diffuse anomalies, like algal blooms or oil spills, posed annotation challenges, occasionally leading to false positives or negatives. This dataset was used to evaluate model performance and to train the multi-class NLSVM for anomaly characterization, with an anomaly subset integrated into the sea/no-sea NLSVM training and a sea patches subset into the anomaly detector training.

### 3.2. Φsat-2 and IMAGIN-e Simulators

To generate images representative of the Φsat-2 and IMAGIN-e missions, we used ESA-provided simulators based on the EO-learn library. These simulators generate images from Sentinel-2 Level 1C data by considering sensor characteristics such as spatial resolution, spectral band shifts, signal-to-noise ratio, and the modulation transfer function. The images, saved in GeoTIFF format, can either be Level 1A (top of atmosphere radiance without spatial correction between spectral bands) or Level 1C (top of atmosphere reflectance with co-registered spectral bands). We exclusively used Level 1C images to avoid errors from band misregistration and allow increased algorithm robustness, leveraging availability of Level 1C pre-processing on board the two missions.

## 4. EXPERIMENTS

This section outlines the model training process and presents the detection results for anomalies (oil spills, algal blooms, and sediments) using simulated images from the Φsat-2 and IMAGIN-e missions. We also evaluate the pipeline's performance on various hardware platforms in terms of latency and throughput.

### 4.1. Settings of Model Trainings

To train the sea/no-sea NLSVM, we applied a weighted binary cross-entropy loss to give more importance to false positives, reducing the chances of incorrectly retaining land patches during anomaly detection. For the multi-class NLSVM, we used a cross-entropy loss.
The encoder was trained using a normalized temperature cross-entropy loss [15], allowing unsupervised training on patches from the normality dataset. After testing different output dimensions, we settled on an architecture that compresses input patches (32 × 32 pixels, 7 bands for IMAGIN-e, 8 bands for Φsat-2) into a 16-dimensional vector. Increasing the dimensionality did not significantly improve performances.
Finally, the GMM was trained on normal sea patches from both the normality and anomaly datasets. Using the Expectation-Maximization (EM) algorithm in scikit-learn [30], we optimized the Gaussian components and their numbers to model the distribution of these patches.

### 4.2. Algorithmic Results

We first evaluate the performance of the pipeline by analyzing each model's effectiveness on representative images of anomalies from both the IMAGIN-e and Φsat-2 missions. We then examine the latent space generated by the encoder to better understand how different anomalies are detected. Finally, we compare the performance of various anomaly detectors for the IMAGIN-e encoder based on the anomaly type.

#### 4.2.1. *Overall Performance of the Pipeline*

The pipeline’s inference results on the three types of anomalies (oil spills, algal blooms, and sediments) are shown in Fig. 2. For each input image, it displays the output of the three stages: sea segmentation, anomaly detection, and anomaly characterization, along with errors in terms of true positives, false positives, and false negatives. Sea segmentation closely matches the ground truth, and anomaly characterization provides good classification accuracy for most types. However, false positives in anomaly detection are mainly due to annotation difficulties that do not fully capture the anomaly areas. While the network, trained only on normal sea data, successfully distinguishes sea from algae and sediment anomalies, it struggles to detect oil spills. Further analysis of the latent space (see next section) explains this limitation.

#### *4.2.2. Latent Space Analysis*

To understand the encoder’s output, we used pixel-level masks to classify patches into categories like sea, land, cloud, and anomalies (oil spill, algal bloom, sediments). We reduced the 16-dimensional latent space using Principal Component Analysis (PCA) (Fig. 3), with the three main PCA axes explaining over 70% of the variance. The SimCLR training, although unsupervised, effectively separates different patch classes into clusters. Mixed classes like coastlines (sea + land) fall between the sea and land clusters.
Although anomalies are close to the sea cluster, algae and sediment patches show clear separation from normal sea patches in the latent space, while oil spill patches remain closer to the sea cluster and are harder to distinguish. In the following section, we will observe that the GMM is capable of correctly identifying algae and sediment anomalies, although oil spills remain more challenging to detect.

#### *4.2.3. Evaluation of Different Anomaly Detection Methods*

Based on the encoded space analyzed previously, we compared the performance of various classical machine learning techniques implemented in the scikit-learn library. We evaluated the anomaly dataset in terms of precision, recall, and F1-score by measuring the separation between normal sea patches and the three anomaly types. The results, shown in Table 1, indicate that detection performance varies by anomaly type.

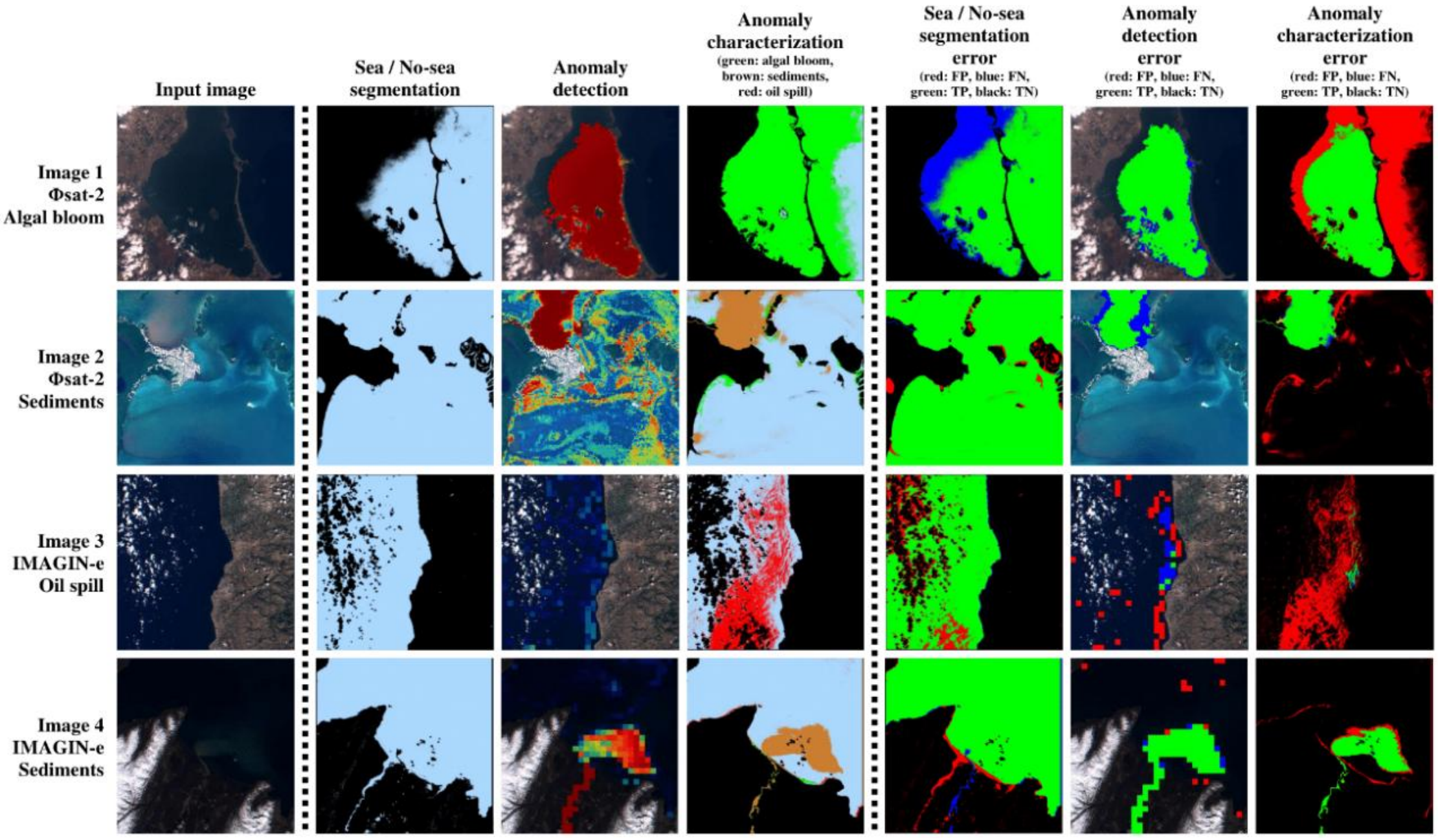

Fig. 2. Example of model prediction on four images. Column 1 shows the images, columns 2–4 show the reconstructed output of the different models, columns 5–7 show the prediction errors of these models compared to the ground truth.

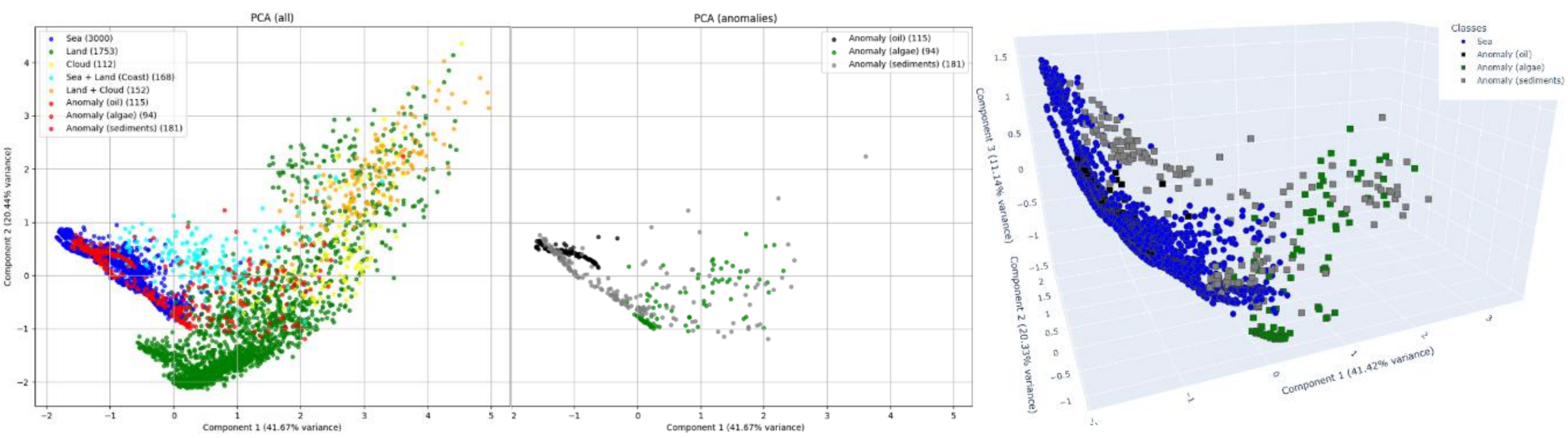

Fig. 3. PCA analysis of labeled patches in the encoder's latent space. The left figure shows the separation of different patch classes (anomalies in red), the middle figure colors anomalies by type, and the 3D PCA on the right illustrates the separation between sea and anomalies. Patch counts per label are indicated in the legend.

Table 1. Performance comparison of machine learning models for detecting algal blooms, sediments, and oil spills.

| | Algal blooms | | | Sediments | | | Oil spills | | |
|---|---|---|---|---|---|---|---|---|---|
| **Model** | **Precision** | **Recall** | **F1-score** | **Precision** | **Recall** | **F1-score** | **Precision** | **Recall** | **F1-score** |
| **GMM** | 95.3% | 87.2% | 91.1% | 88.5% | 76.2% | 81.9% | 14.3% | 1.7% | 3.1% |
| **Local Outlier Factor** | 85.8% | 96.8% | 91.0% | 67.8% | 43.1% | 52.7% | 13.0% | 6.1% | 8.3% |
| **Isolation Forest** | 56.4% | 98.9% | 71.8% | 87.5% | 42.5% | 57.2% | 0.0% | 0.0% | 0.0% |
| **One Class SVM** | 47.8% | 94.7% | 63.6% | 94.2% | 35.9% | 52.0% | 0.0% | 0.0% | 0.0% |
| **Elliptic Envelope** | 29.6% | 88.3% | 44.4% | 49.2% | 49.2% | 49.2% | 1.8% | 1.7% | 1.8% |

For algae and sediment anomalies, the GMM performed best, effectively modeling normal sea conditions with around twenty Gaussian components, leading to superior detection accuracy. In contrast, methods like Local Outlier Factor (LOF) [31], One-Class SVM (OCSVM) [32], and Isolation Forest (IF) [33] delivered moderate performance on average. LOF, which detects outliers based on local density, works well in certain cases but struggles with sparse or poorly separated data. OCSVM creates a hyperplane to distinguish normal data from anomalies but is less effective in complex distributions. IF uses decision trees to isolate anomalies but performs less consistently in capturing intricate normality patterns. Elliptic Envelope (EE) [34] performs poorly, as it models normality with a single Gaussian distribution. This approach is too simplistic to handle the complexity of the data, especially for multiple anomaly types, resulting in poor separation between normal and anomalous patches and subpar detection performance.

### 4.3. Hardware Results

We deployed our models on hardware platforms representing the onboard systems of the Φsat-2 and IMAGIN-e missions. For Φsat-2, the patch encoder was deployed on the Myriad™ 2 accelerator, while the NLSVM models and the GMM were run on an ARM Cortex-A53 processor. For IMAGIN-e, which lacks a dedicated hardware accelerator, we used an ARM Cortex-A72 CPU for deployment and evaluation. Additionally, we tested the encoder on both the Myriad™ 2 and the newer Myriad™ X to assess the impact of different accelerators.

Performance results (see Table 2) show that NLSVM models are lightweight and run efficiently on both ARM CPUs, with especially high throughput on the Cortex-A72. The GMM performs even faster—about ten times quicker than the NLSVMs—due to operating on encoded patches with only 16 values instead of full 32 × 32 pixel patches, leading to faster anomaly detection. While the encoder is relatively small, with one million parameters, it remains the bottleneck in the pipeline, taking the most processing time. On the ARM A72 CPU, its throughput significantly decreases without an accelerator. However, using hardware acceleration boosts performance substantially, improving throughput by three times on Myriad™ 2 and nine times on Myriad™ X.

In summary, the hardware platforms chosen reflect the computational limitations of Φsat-2 and IMAGIN-e. While CPU-based models like NLSVM run efficiently, neural network tasks like the encoder benefit greatly from hardware accelerators. The GMM's efficiency, operating on compressed data, makes it a standout for fast anomaly detection.

## 5. DISCUSSION

Our study presents a practical and adaptable solution for detecting a wide range of anomalies in marine ecosystems using embedded AI on Earth observation satellites. By prioritizing the transmission of relevant data and generating alerts, our approach ensures that only critical information is sent to the ground. This selective transmission not only optimizes limited satellite communication bandwidth but also improves system responsiveness in monitoring vast oceanic areas.

A key advantage of our method is the significant data reduction achieved by transmitting only anomalous or high-priority data. By identifying areas of interest, our onboard algorithms efficiently reduce the volume of information sent, allowing for early alerting and low-bandwidth monitoring. In addition, we explored working in a 16-dimensional latent space, where each input image patch (32 × 32 pixels with 8 spectral bands) is compressed to just 16 values, achieving a compression factor of over 500. While this drastically reduces data flow, it also opens possibilities for more advanced processing techniques, such as anomaly detection, change detection, and even image reconstruction. These approaches could leverage the latent space information for more complex ground-based analyses, optimizing the balance between onboard and post-processing tasks.

Furthermore, considering the hypothetical mission configuration of a LEO satellite in Table 3, the satellite needs to process 152 patches per second to analyse sensed images on-the-flow. As shown in Table 2, this processing rate is achievable with the MyriadX accelerator, enabling real-time onboard analysis of large areas like the ocean. This capability opens the door to continuous background processing over extensive regions, allowing real-time detection and transmission of compressed information for further onboard or ground-based analysis.

While our solution shows strong potential, certain limitations remain. Detecting some anomalies, such as oil spills, is challenging due to insufficient separation in the latent space. Improving the latent space representation, possibly through weakly supervised methods, could enhance detection performance for these difficult cases.

Deploying our solution on actual missions like Φsat-2 and IMAGIN-e will allow us to validate its operational effectiveness with real in-orbit data. This will help thoroughly evaluate system responsiveness and robustness under diverse environmental conditions, guiding future improvements.

In conclusion, our approach not only advances onboard AI capabilities but also provides a scalable framework for future Earth observation missions. By focusing on selective transmission, we optimize satellite and ground resources, ensuring timely and efficient responses to environmental threats.

Table 2. Latency and throughput comparison of AI models for patch processing across different deployment targets.

| AI Model | Target | Latency (µs/patch) | Throughput (patches/s) |
|---|---|---|---|
| Sea/No-Sea NLSVM | ARM A53 | 202 | 4,955 |
| | ARM A72 | 89 | 11,198 |
| Encoder (128 batch size) | Myriad™ 2 | 16,583 | 60 |
| | Myriad™ X | 5,703 | 175 |
| | ARM A72 | 57,421 | 17 |
| GMM | ARM A53 | 29.3 | 34,129 |
| | ARM A72 | 16 | 63,694 |
| Multi-class NLSVM | ARM A53 | 1,125 | 888 |
| | ARM A72 | 514 | 1,946 |

Table 3. Processing capabilities for LEO satellite imaging.

| **Ground Speed** | **GSD (resolution)** | **Throughput (speed dir.)** | **Swath** | | **Throughput** | | **Data rate (FL16 bits)** |
|---|---|---|---|---|---|---|---|
| 7,778 m/s | 50 m/pixel | 155.56 pixels/s | 50 km | 1,000 pixels | 156K pixels/s | 152 patches/s | 304 bytes/s |

## 6. CONCLUSION

We introduced a practical and adaptable solution for onboard anomaly detection in marine ecosystems using embedded AI on Earth observation satellites. Our approach combines a self-supervised neural network encoder with a Gaussian Mixture Model to detect anomalies as deviations from normal sea conditions. The encoder reduces input data from 8,192 values to just 16, simplifying onboard processing and enabling the extraction of useful and robust information from the input image, which is crucial for resource-limited satellites.
Our experiments showed effective detection of anomalies like algal blooms and sediments, but also highlighted challenges in detecting oil spills due to overlap in the latent space. Deploying our models on hardware representative of the Φsat-2 and IMAGIN-e missions confirmed their suitability for resource-constrained environments, particularly when using hardware accelerators.
By prioritizing critical information and reducing the volume of transmitted data, our solution enhances satellite responsiveness in monitoring vast oceanic areas. This approach not only optimizes satellite communication bandwidth but also opens up possibilities for advanced ground-based analysis using latent space representations, enabling more complex processing such as anomaly detection, change detection, or image reconstruction.
In summary, our method advances onboard AI capabilities, contributes to efficient data reduction in satellite missions, and provides a scalable framework for future marine environmental monitoring applications, helping to protect ocean ecosystems more effectively.

## ACKNOWLEDGMENTS

The work presented in this paper is part of the IRMA project (Image processing for a Responsive Mission with AI) led by IRT Saint Exupéry. The project focuses on developing AI-based technologies for mission planning and data processing, both onboard and on the ground, to meet the evolving needs of the Earth observation satellite market, including data interpretation and the reactive management of satellite constellations. IRMA also aims to demonstrate the effectiveness of these technologies through end-to-end system demonstrations, in both lab and orbital environments.
The authors would like to thank the industrial and academic partners: Thales Alenia Space, Activeeon, Geo4i, JoliBrain, and the University Côte d'Azur. Special thanks to Φ-lab at European Space Agency for enabling the deployment of these technologies onboard the Φsat-2 and IMAGIN-e missions.